\documentclass{article}
\PassOptionsToPackage{numbers, compress}{natbib}
\usepackage[final,main]{neurips_2026}

\usepackage[utf8]{inputenc}
\usepackage[T1]{fontenc}
\usepackage[hidelinks]{hyperref}
\usepackage{url}
\usepackage{booktabs}
\usepackage{amsfonts}
\usepackage{amsmath}
\usepackage{amssymb}
\usepackage{mathtools}
\usepackage{microtype}
\usepackage{xcolor}
\usepackage[table]{xcolor}
\usepackage{graphicx}
\usepackage{algorithm}
\usepackage{algpseudocode}
\usepackage{wrapfig}
\usepackage{booktabs}
\usepackage{multirow}
\usepackage{caption}
\usepackage{fvextra}

\title{Learning Where to Look: Observation Policy Optimization for Thinking with Images}

\author{
  Junfeng Wang$^{1,*}$ \quad Jiawei Liu$^{1,*}$ \quad Yongchao Xu$^{1,*}$ \quad Tao Jiang$^{2}$ \\
  \bfseries Jiangbo Ai$^{2}$ \quad Jin Zhang$^{2}$ \quad Chong Wang$^{1,\dagger}$ \quad Qixing Zhang$^{1}$ \\[4pt]
  \normalfont $^{1}$University of Science and Technology of China \\
  \normalfont $^{2}$Alibaba Group \\[3pt]
  \normalfont\small \texttt{\{jfwang2002,yongchaoxu\}@mail.ustc.edu.cn} \\
  \normalfont\small \texttt{\{jwliu6,wchong,qixing\}@ustc.edu.cn} \\
  \normalfont\small \texttt{\{cigarette.jt,aijiangbo.ajb,zj146372\}@alibaba-inc.com} \\[3pt]
  \normalfont\small $^{*}$Equal contribution.\quad $^{\dagger}$Corresponding author.
}

\newcommand{\method}{OPO}
\DefineVerbatimEnvironment{promptbox}{Verbatim}{
  fontsize=\scriptsize,
  frame=single,
  framesep=2mm,
  breaklines=true,
  breakanywhere=true,
  breaksymbolleft={},
  breaksymbolright={}
}

\begin{document}

\maketitle
\begin{abstract}
Complex visual question answering often requires vision-language models to \emph{think with images}, i.e., actively acquire local visual evidence by zooming into informative regions of high-resolution images. Our error analysis shows that many failures in this setting originate from incorrect observation actions, where the model fails to localize key regions or obtain crops that contain sufficient evidence for answering. We further observe that bounding-box tokens exhibit substantially higher generation entropy than ordinary language tokens, indicating high uncertainty when the model decides where to look. However, existing reinforcement learning methods based on final-answer rewards provide only trajectory-level feedback, making it difficult to assign precise credit to the observation actions.
We propose Observation Policy Optimization (OPO), an RL algorithm for optimizing observation actions in tool-augmented thinking with images. OPO consists of two core modules. First, Uncertainty-Guided Observation Branching (UOB) treats each image zoom-in tool call as an observation event and selectively branches at high-uncertainty, low-redundancy events during rollout, enabling efficient exploration of alternative cropping decisions and their resulting visual evidence. Second, Evidence-Localized Advantage Attribution (ELAA) compares sibling branches using local evidence-sufficiency rewards and assigns credit only to the corresponding observation-action tokens. By decoupling observation-action optimization from trajectory-level outcome supervision, OPO improves visual evidence acquisition and further enhances visual reasoning performance. Experiments show that OPO consistently improves Qwen3-VL models across scales and achieves competitive performance against larger models and tool-augmented baselines.
Code is publicly available.\footnote{\url{https://github.com/Yann0122/Observation-Policy-Optimization}}
\end{abstract}

\section{Introduction}

Vision-language models (VLMs) have made substantial progress in visual question answering, chart understanding, document parsing, and complex scene reasoning~\citep{bai2025qwen25vltechnicalreport, bai2025qwen3, zhu2025internvl3, team2025kimi, guo2025seed1}. However, many high-resolution visual tasks cannot be solved from a single global image view. The evidence required to answer a question may appear only in small text, subtle objects, spatial relations, or local image regions. In such cases, models must actively acquire local visual evidence by selecting informative regions and zooming in, an ability often referred to as \emph{thinking with images}~\citep{chu2025qwen, zheng2025deepeyes, fan2025grit, cao2025ground, su2025thinkingimagesmultimodalreasoning}.

\begin{figure}[t]
    \centering
    \includegraphics[width=\linewidth]{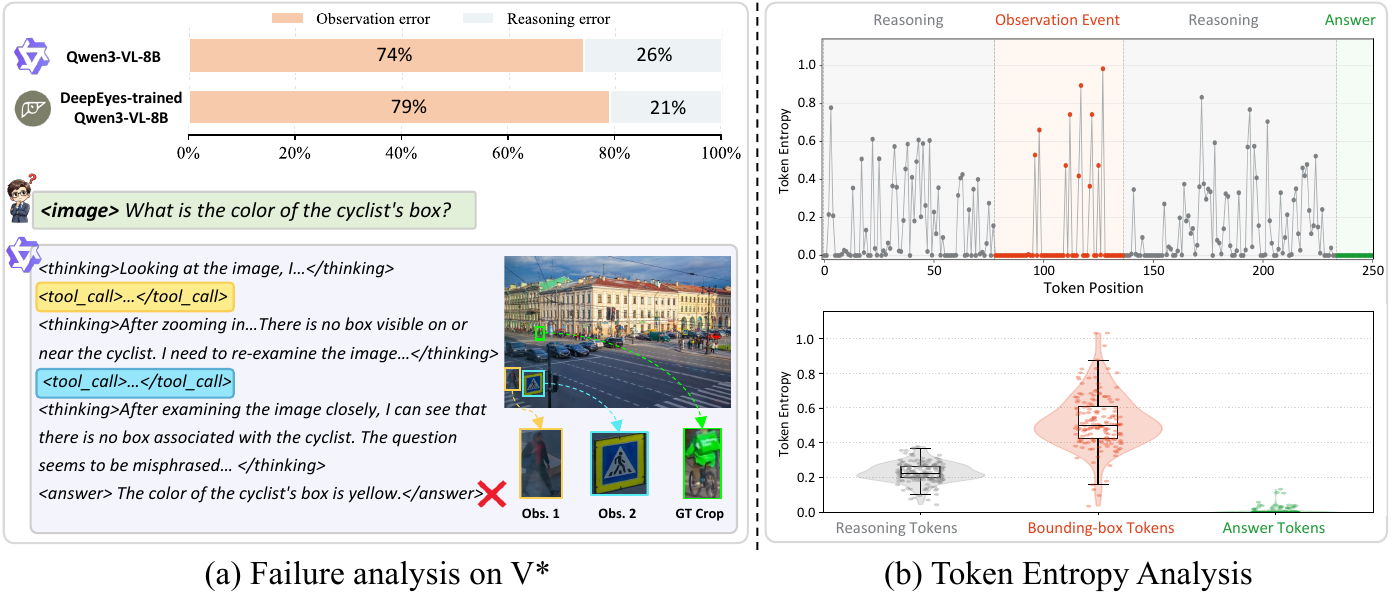}
    \caption{Motivation for observation-action optimization. 
(a) Error-type breakdown of incorrect V* trajectories for Qwen3-VL-8B and DeepEyes-trained Qwen3-VL-8B, together with an example observation-error trajectory. 
(b) Token-entropy analysis, where the top panel shows an event-aligned entropy trace over a generated response and the bottom panel compares entropy distributions of reasoning tokens, bounding-box tokens, and answer tokens.}
    \label{fig:motivation}
\end{figure}

In this paper, we study tool-augmented thinking with images, where a VLM invokes an image zoom-in tool during multi-turn reasoning. In this setting, the key challenge is not only to reason over visual evidence, but also to acquire the right evidence in the first place. As shown in Figure~\ref{fig:motivation}(a), our error analysis shows that many failures stem from incorrect observation actions: the model either localizes the wrong region or obtains crops that lack sufficient evidence. We further find that bounding-box tokens exhibit substantially higher generation entropy than ordinary language tokens, as shown in Figure~\ref{fig:motivation}(b), suggesting that visual localization is a high-uncertainty local decision.

Existing RL-based methods for visual tool use typically optimize tool-augmented reasoning with trajectory-level rewards, such as final-answer correctness, format validity, or tool-use rewards~\citep{zheng2025deepeyes, fan2025grit, cao2025ground, shao2024deepseekmath}. While such rewards can encourage models to use visual tools, they provide imprecise credit assignment for observation actions. A crop that contains sufficient evidence may be penalized if later reasoning fails, while an insufficient crop may be reinforced if the model guesses the answer correctly. This credit mismatch makes it difficult for the model to reliably learn where to look.

To address this problem, we propose Observation Policy Optimization (OPO), a reinforcement learning algorithm that directly optimizes observation actions in tool-augmented thinking with images. OPO consists of two complementary modules. First, Uncertainty-Guided Observation Branching (UOB) treats each zoom-in tool call as an observation event and selectively branches from high-uncertainty, low-redundancy events, allowing the model to explore alternative crops from a shared reasoning prefix. Second, Evidence-Localized Advantage Attribution (ELAA) compares sibling branches using local evidence-sufficiency rewards and assigns the resulting advantage only to the observation-action tokens that produced the corresponding crop. Together, UOB and ELAA separate observation-action optimization from trajectory-level outcome supervision, enabling more precise learning of where to look.

We evaluate OPO using Qwen3-VL backbones~\citep{bai2025qwen3} on V*~\citep{wu2024v}, HR-Bench 4K, and HR-Bench 8K~\citep{wang2025divide}. OPO consistently improves both model scales, increasing overall accuracy by 10.0, 12.4, and 15.8 percentage points for Qwen3-VL-2B and by 4.7, 4.1, and 4.3 percentage points for Qwen3-VL-8B on the three benchmarks, respectively. OPO-8B also achieves competitive performance against strong open-source, proprietary, and tool-augmented baselines across these benchmarks. Ablation studies further show that uncertainty-guided observation branching and evidence-localized advantage attribution both contribute to these gains.

Our contributions are summarized as follows:
\begin{itemize}
    \item We identify observation actions as a key bottleneck in thinking with images. Our analysis shows that many failures stem from insufficient visual evidence acquisition, and that observation-action tokens exhibit high local uncertainty.

    \item We propose OPO, which consists of two complementary modules: Uncertainty-Guided Observation Branching for targeted exploration of alternative cropping decisions, and Evidence-Localized Advantage Attribution for precise credit assignment.

    \item Experiments show that OPO consistently improves Qwen3-VL models across scales and achieves competitive performance against much larger models and tool-augmented methods. 
\end{itemize}

\section{Related Work}
\label{app:related-work}

\paragraph{Thinking with Images.}
Thinking with images extends multimodal reasoning by allowing models to acquire and manipulate visual evidence during inference~\citep{su2025thinkingimagesmultimodalreasoning}. Within this paradigm, visual exploration methods use crops, zooming, and other image operations to inspect details that may be lost in the global view~\citep{zhang2025chain, zhu2025active, su2505openthinkimg}. DeepEyes learns a zoom-based observation policy through reinforcement learning~\citep{zheng2025deepeyes}. Broader tool interfaces support more diverse evidence acquisition: Thyme uses executable code for image manipulation and computation~\citep{zhang2025thyme}, while DeepEyesV2 and SenseNova-MARS combine visual operations with search~\citep{hong2025deepeyesv2, chng2025sensenovamars}. Alongside these interactive approaches, latent visual reasoning explores internal visual representations~\citep{li2025latent}, and ZwZ distills region-level supervision into the model to enable single-pass perception without inference-time zooming~\citep{wei2026zwz}.

A related question is how to supervise the visual actions taken during reasoning. CodeVision and CodeV address this through process rewards for code-based tool use. CodeVision rewards tool-use strategy and efficiency, including improvements in crop overlap with a target region~\citep{guo2025codevision}. CodeV evaluates the evidence provided by tool outputs and combines these scores with answer correctness to form a trajectory-level advantage~\citep{hou2025codev}. OPO connects evidence supervision to the observation decision itself: it explores alternative crops from a shared reasoning prefix, compares their evidence sufficiency within the resulting sibling group, and assigns the local advantage to the corresponding observation-action tokens. This couples the exploration of candidate observations with credit assignment for the evidence they acquire.

\paragraph{Agentic Reinforcement Learning.}
Agentic reinforcement learning formulates large language models as learnable policies embedded in multi-turn interactive environments, enabling them to plan, reason, and invoke tools under partial observations~\citep{zhang2025landscape}. More broadly, reinforcement learning has also found applications in other domains~\citep{ye2026coevomem, xu2026videoevolve, xu2026diversifyfocus}. Early tool-augmented agents mainly relied on prompting or supervised fine-tuning to elicit tool-use behaviors, such as ReAct, Toolformer, FireAct, AgentTuning, and Agent-FLAN~\citep{yao2022react, schick2023toolformer, chen2023fireact, zeng2310agenttuning, chen2024agent}. Recent studies further apply reinforcement learning to multi-turn tool interaction, using outcome or rule-based rewards to train models to autonomously invoke search engines, browsers, code interpreters, or visual tools for complex reasoning, information seeking, and multimodal tasks~\citep{qian2025toolrlrewardtoollearning, feng2025retool, wei2507autotir, jin2025search, wu2025webdancer, li2025webthinker, li2025websailor, wu2025vtool, zheng2025deepeyes, wang2025pixel}. ARPO and AEPO further leverage token entropy to improve multi-turn tool-use exploration through branch sampling or entropy-balanced optimization at high-uncertainty tool-call steps~\citep{dong2025arpo, dong2025aepo}. While these methods target general tool-use behavior, OPO specializes agentic RL to visual observation: it uses uncertainty-guided branching to explore alternative crops and evidence-localized attribution to assign feedback according to the visual evidence produced by the observation action.

\section{Preliminary Analysis}

\subsection{Observation Events and Observation-Action Tokens}

We follow the image zoom-in tool interface of Qwen3-VL~\citep{bai2025qwen3}, where the model generates tool-call tokens in a predefined format to trigger a zoom-in operation. Each tool call contains a region label, a bounding box, and an image index, which together determine the local visual evidence the model will observe next. We refer to each valid image zoom-in tool call as an \emph{observation event}, and call its region-label, bounding-box and image-index tokens \emph{observation-action tokens}. The concrete tool-call schema, prompt template, and zoom-in tool behavior are provided in Appendix~\ref{app:prompt-tool-configuration}.

\subsection{Failure Analysis on V*}

We analyze the error trajectories of Qwen3-VL-8B-Instruct~\citep{bai2025qwen3} and a Qwen3-VL-8B model trained under the DeepEyes~\citep{zheng2025deepeyes} paradigm on the V*~\citep{wu2024v} dataset, aiming to understand the main sources of failure in tool-augmented visual reasoning. We focus on whether the model's image zoom-in operation acquires local visual evidence sufficient to answer the question. Specifically, we categorize a trajectory as an observation error if the model localizes an irrelevant region that does not contain the key evidence required for answering the question. If the cropped region contains sufficient evidence but the model still fails in subsequent reasoning or answer generation, we categorize it as a reasoning error.
As shown in Figure~\ref{fig:motivation}(a, top), Qwen3-VL-8B-Instruct achieves an answer accuracy of 87.96\% on V*. Among its incorrect answers, approximately 74\% of failures are mainly caused by observation errors. For the Qwen3-VL-8B model trained under the DeepEyes paradigm, the answer accuracy improves to 90.05\%, yet approximately 79\% of its incorrect answers still mainly originate from observation errors.
These results indicate that although outcome-level RL can improve final-answer accuracy, failures remain concentrated in the local visual evidence acquisition stage. Therefore, the training process needs to provide more direct optimization signals for the observation actions that determine where the model looks. Localization failures also dominate unsuccessful trajectories in stronger models (Appendix~\ref{app:stronger-model-localization}).

\subsection{Token Entropy Analysis}

We further examine the token-level generation entropy of different parts of the model response. For decoding position $t$, let $p_t$ denote the next-token distribution over the full vocabulary, conditioned on the input question $x$, the tool context $\mathcal{T}$, and the previously generated prefix $\mathcal{R}_{<t}$. We define the token entropy at this position as
\begin{equation}
H_t = - \sum_{j=1}^{V} p_{t,j}\log p_{t,j},
\quad
p_t=\pi_\theta(\cdot\mid \mathcal{R}_{<t},x;\mathcal{T})
=\mathrm{Softmax}\!\left(\frac{z_t}{\tau}\right).
\label{eq:token-entropy}
\end{equation}
Here $V$ is the vocabulary size, $p_{t,j}$ is the probability of the $j$-th vocabulary token at position $t$, $z_t\in\mathbb{R}^{V}$ denotes the pre-softmax logits, and $\tau$ is the decoding temperature. This entropy quantifies the uncertainty of the model's next-token prediction at decoding position $t$. For analysis, we group generated tokens into reasoning tokens, observation-event tokens, and answer tokens. Observation-event tokens correspond to the structured image zoom-in tool call.

As shown in Figure~\ref{fig:motivation}(b, bottom), the bounding-box tokens used for localization in observation actions exhibit substantially higher entropy than ordinary reasoning tokens. We conduct this analysis over all samples in the V*~\citep{wu2024v} dataset. The bounding-box tokens have a mean entropy of 0.522 and a median entropy of 0.501, both more than twice the corresponding mean entropy of 0.229 and median entropy of 0.225 for reasoning tokens. In contrast, answer tokens are nearly deterministic, with a mean entropy of only 0.006. The event-aligned trace in Figure~\ref{fig:motivation}(b, top) further shows that entropy spikes concentrate on bounding-box tokens, which directly determine where the model looks. This observation aligns with entropy-guided agentic RL, where high-entropy tool-use steps are treated as promising branching points for adaptive exploration~\citep{dong2025arpo}. These results indicate that visual localization is a high-uncertainty observation decision, motivating uncertainty-guided branching around observation events.

\section{Method}

Observation Policy Optimization (OPO) is a reinforcement learning framework that improves where a model looks during tool-augmented thinking with images. Following the definition above, we denote each candidate observation event by $e$. \method\ introduces two complementary modules for optimizing the model's observation actions: Uncertainty-Guided Observation Branching (UOB, Section~\ref{sec:uob}) and Evidence-Localized Advantage Attribution (ELAA, Section~\ref{sec:elaa}). The full training procedure is summarized in Appendix~\ref{app:algorithm-procedure}.

\begin{figure}[t]
    \centering
    \includegraphics[width=\linewidth]{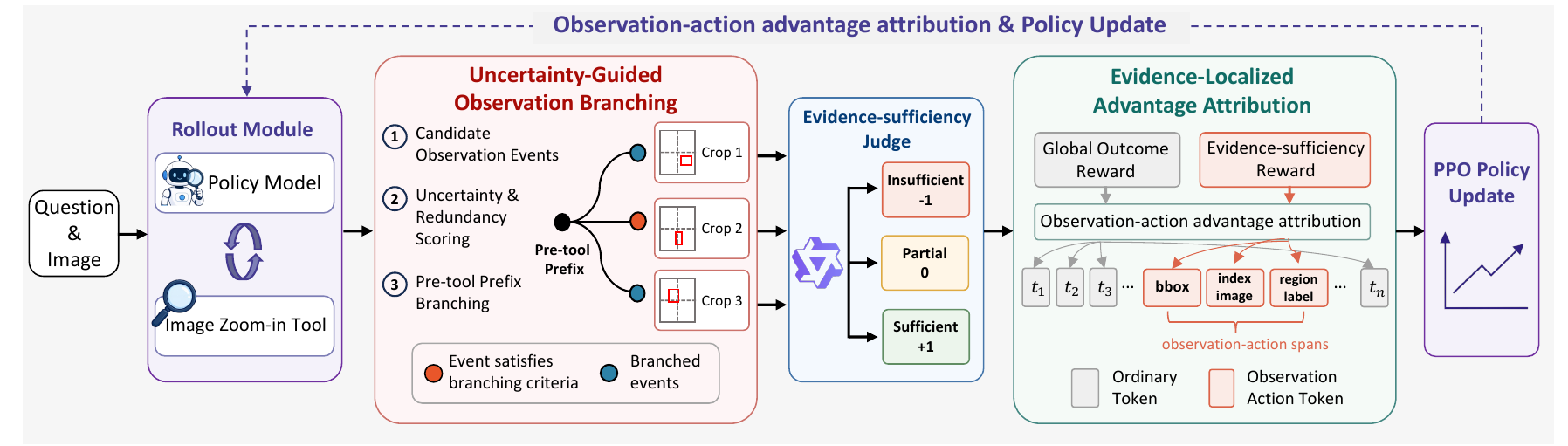}
    \caption{Overview of \method. UOB concentrates rollout exploration around high-uncertainty and low-redundancy observation decisions by branching from the shared pre-tool prefix. ELAA then compares sibling crops within each branch group using evidence-sufficiency rewards and assigns the resulting local advantage only to the observation-action tokens.}
    \label{fig:algorithm}
\vspace{-0.9em}
\end{figure}

\subsection{Uncertainty-Guided Observation Branching}
\label{sec:uob}

UOB is a rollout-time module for targeted exploration around uncertain observation events. Instead of spending all $M$ samples on independent full-trajectory rollouts, UOB first obtains a small set of base trajectories and then uses the remaining budget to branch at high-uncertainty observation events while filtering redundant child candidates. This preserves the same reasoning prefix while varying the observation action, making the resulting sibling trajectories directly comparable. As a result, UOB concentrates exploration on observation decisions where the model is most unsure, avoids spending rollout budget on redundant crops, and produces branch groups that provide controlled comparisons for ELAA.

\paragraph{Candidate observation events.}

For each question, \method\ produces $M$ trajectories under a fixed rollout budget. It first samples $B$ base trajectories and reserves the remaining $P=M-B$ rollout slots for observation-event branching and backfill. Each base trajectory is decoded until it reaches the first valid observation event or terminates without invoking the zoom-in tool. For each collected event, we record the shared pre-tool prefix, the generated observation-action spans, and the original candidate crop. These event records provide the basis for UOB to branch from the shared pre-tool prefix and for ELAA to attribute local feedback to the corresponding observation-action tokens.

\paragraph{Event uncertainty.}

UOB scores each candidate event by the local token entropy around the observation action. We use $H_t$ defined in Eq.~\eqref{eq:token-entropy} as the uncertainty at decoding position $t$. A larger $H_t$ indicates a more diffuse next-token distribution and thus higher uncertainty.

Motivated by the entropy analysis above, UOB localizes event uncertainty to bounding-box generation. Let $S_{\mathrm{box}}(e)$ denote the bounding-box token span of event $e$. Since the bounding box specifies the local region to be observed, its generation entropy provides a direct signal of uncertainty in the observation decision.

\begin{equation}
 g(e)=
 \mathrm{mean}_{t\in S_{\mathrm{box}}(e)}H_t
 + \lambda_{\mathrm{std}}\,\mathrm{std}_{t\in S_{\mathrm{box}}(e)}H_t.
\end{equation}
The standard-deviation term makes the score sensitive to entropy spikes within the box span, which often indicate unstable coordinate or region decisions; $\lambda_{\mathrm{std}}$ controls the strength of this spike-sensitive signal.
A larger $g(e)$ indicates that the event lies in a more uncertain localization decision region.

\paragraph{Question-adaptive anomaly and redundancy.}

Because raw uncertainty varies across questions and prompts, UOB normalizes event uncertainty against the question's own local baseline. Let $b(e)$ be the mean uncertainty in a fixed window of $W$ tokens before the bounding-box span. Over the base candidates $\mathcal{E}_x$ of the same question, we compute
\begin{equation}
 z(e)=\frac{g(e)-\mathrm{mean}_{e'\in\mathcal{E}_x}b(e')}
 {\mathrm{std}_{e'\in\mathcal{E}_x}b(e')+\epsilon}.
\end{equation}
Here $\mathcal{E}_x$ is the set of base candidate events for question $x$, $b(e')$ is the pre-action baseline uncertainty of candidate event $e'$, and $\epsilon$ is a numerical stability constant. The resulting $z(e)$ is a question-normalized anomaly score: it measures how much the bbox-based event uncertainty $g(e)$ deviates from the typical pre-action uncertainty for the same question.
In this form, a large $z(e)$ means that the observation event is unusually uncertain relative to the model's local pre-action uncertainty on the same question.

To avoid repeatedly exploring similar regions within the same branch group, UOB computes a crop redundancy score for each child candidate $c$ sampled from event $e$:
\begin{equation}
\mathrm{Red}(c\mid e) = \max_{c'\in\mathcal{C}_e} \mathrm{IoU}(b_c,b_{c'}),
\end{equation}
where $b_c$ is the bounding box of candidate $c$, and $\mathcal{C}_e$ contains the original candidate and previously accepted child crops on the same image within the branch group. UOB filters out redundant child candidates, allocating the limited rollout budget to diverse observation regions.

\paragraph{Branching from the observation prefix.}

The branching rule combines uncertainty-guided budget allocation with redundancy-based candidate filtering:
\begin{equation}
Z(e)=\mathcal{A}\!\left(z(e), P_{\mathrm{rem}}\right),
\end{equation}
where $\mathcal{A}$ is the budget allocator, $Z(e)$ is the maximum number of child rollouts proposed for event $e$, and $P_{\mathrm{rem}}$ is the remaining branch budget. The allocator assigns more proposals to high-uncertainty events and always respects the remaining rollout budget and the number of available trajectory slots. Redundancy filtering determines which proposals are retained.

A child trajectory is not generated from scratch. Instead, \method\ preserves the prompt and all tokens up to the pre-tool prefix boundary, then resamples the suffix from that prefix. The newly generated suffix may contain an alternative observation event. Valid child tool calls are accepted if their redundancy scores do not exceed a threshold $\tau_{\mathrm{red}}$. Accepted candidates are added to $\mathcal{C}_e$, their tool calls are executed, and the agent continues multi-turn generation until termination. If fewer than $M$ trajectories are obtained after all branch opportunities, ordinary rollouts are used to backfill the group. The trajectories derived from the same pre-tool prefix form a branch group, which serves as the comparison set for ELAA.

\subsection{Evidence-Localized Advantage Attribution}
\label{sec:elaa}

ELAA converts the sibling trajectories produced by UOB into local advantages for observation-action tokens. Trajectories in the same branch group share the same pre-tool prefix, but may generate different observation actions and crops. ELAA first evaluates each crop with an evidence-sufficiency reward, then normalizes these local scores within the branch group to obtain a group-local advantage. Finally, this local advantage is applied only to the observation-action token spans that produced the corresponding crop, while the remaining generated tokens keep the global outcome advantage.

\paragraph{Global outcome reward.}

The global outcome reward is computed from the final answer and task-level formatting and tool-use signals:
\begin{equation}
R_i^{\mathrm{global}}
= w_{\mathrm{acc}}r_i^{\mathrm{acc}}
+ w_{\mathrm{fmt}}r_i^{\mathrm{fmt}}
+ w_{\mathrm{tool}}r_i^{\mathrm{tool}}.
\end{equation}
Here $R_i^{\mathrm{global}}$ is the trajectory-level outcome reward for trajectory $i$. The terms $r_i^{\mathrm{acc}}$, $r_i^{\mathrm{fmt}}$, and $r_i^{\mathrm{tool}}$ respectively denote answer correctness, format quality, and tool-use quality~\citep{zheng2025deepeyes}. The coefficients $w_{\mathrm{acc}}$, $w_{\mathrm{fmt}}$, and $w_{\mathrm{tool}}$ control the relative weights of these outcome-level signals.

\paragraph{Evidence-sufficiency reward.}

For each evidence crop, a frozen visual judge receives the question, the ground-truth answer, and the crop, and is prompted to assess whether the crop contains sufficient visual evidence to support the answer. The judge labels the crop as \textsc{Insufficient}, \textsc{Partial}, or \textsc{Sufficient}. The reward-judge prompts and model choices are detailed in Appendix~\ref{app:reward-evidence-judge}.
The default scalar mapping is
\begin{equation}
\rho(\textsc{Insufficient})=-1,\quad
\rho(\textsc{Partial})=0,\quad
\rho(\textsc{Sufficient})=1.
\end{equation}
Here $\rho(\cdot)$ maps the judge's discrete evidence-sufficiency label to a scalar local reward: insufficient evidence is penalized, partial evidence is neutral, and sufficient evidence is rewarded.

For each sibling trajectory $i$ in branch group $g$, let $c_{i,g}$ denote the crop produced by its observation action. The group-local evidence reward is
\begin{equation}
R_{i,g}^{\mathrm{evi}}=\rho(c_{i,g}).
\label{eq:local-evidence-reward}
\end{equation}
$R_{i,g}^{\mathrm{evi}}$ measures the quality of visual evidence acquired by the local observation action. This reward is used only for local comparison among sibling trajectories in the same branch group.

\paragraph{Observation-action advantage attribution.}

After rollout and reward computation, standard GRPO first computes a global group advantage using the shared question identifier:
\begin{equation}
A_i^{\mathrm{global}}
=
\frac{R_i^{\mathrm{global}}-\mathrm{mean}_{j\in\mathcal{G}(x)}R_j^{\mathrm{global}}}
{\mathrm{std}_{j\in\mathcal{G}(x)}R_j^{\mathrm{global}}+\epsilon}.
\end{equation}

Here $\mathcal{G}(x)$ is the rollout group for question $x$. The advantage $A_i^{\mathrm{global}}$ indicates whether trajectory $i$ achieves a better or worse outcome than its group peers.

Let $\mathcal{S}(g)$ be the set of sibling trajectories in branch group $g$. ELAA uses only the evidence-sufficiency reward to construct the group-local advantage:
\begin{equation}
A_{i,g}^{\mathrm{local}}
=
\frac{R_{i,g}^{\mathrm{evi}}-\mathrm{mean}_{j\in\mathcal{S}(g)}R_{j,g}^{\mathrm{evi}}}
{\mathrm{std}_{j\in\mathcal{S}(g)}R_{j,g}^{\mathrm{evi}}+\epsilon}.
\end{equation}

Because normalization is performed only within the same branch group, $A_{i,g}^{\mathrm{local}}$ reflects relative evidence-quality differences among alternative observation actions sampled from the same prefix.

Let $m_{i,t}(g)$ be a binary mask that is one only for the image-index, region-label, and bounding-box spans generated by trajectory $i$ in branch group $g$. ELAA applies the evidence-local advantage only to these observation-action tokens:
\begin{equation}
A_{i,t}
=
\left(1-m_{i,t}(g)\right)A_i^{\mathrm{global}}
+ m_{i,t}(g)\lambda_{\mathrm{local}}A_{i,g}^{\mathrm{local}}.
\label{eq:local-global-adv}
\end{equation}

Thus, tokens outside the observation-action spans are trained with the global outcome advantage, while observation-action tokens receive the evidence-local advantage computed within their corresponding branch group. This yields a clean credit assignment between downstream reasoning and visual evidence acquisition.

After obtaining the token-level advantage $A_{i,t}$, \method\ updates the policy with the standard clipped PPO objective. OPO does not change the form of the PPO update; its effect enters through the advantage $A_{i,t}$ defined above. The full policy objective is provided in Appendix~\ref{app:policy-update}.

\section{Experiments}

\subsection{Experimental Setup}
\paragraph{Baselines.}
To comprehensively assess the effectiveness of OPO, we compare it against three categories of baselines. First, we include advanced proprietary models, including GPT-5 mini and GPT-5.2~\citep{openai2025gpt5systemcard,openai2025gpt52systemcard}, Claude Opus 4.6~\citep{anthropic2026claudeopus46}, and Gemini-3 Pro~\citep{googledeepmind2025gemini3pro}, which represent strong closed-source general-purpose vision-language systems. Second, we compare with state-of-the-art open-source vision-language models, including LLaVA-OneVision~\citep{li2024llavaonevisioneasyvisualtask}, Qwen2.5-VL~\citep{bai2025qwen25vltechnicalreport}, Qwen3-VL~\citep{bai2025qwen3}, Qwen3.5~\citep{qwen35blog}, and Kimi K2.5~\citep{kimiteam2026kimik25visualagentic}. Third, we consider methods explicitly designed with workflows or tool-use procedures, including SEAL~\citep{wu2024v}, DyFo~\citep{li2025dyfo}, ZoomEye~\citep{shen-etal-2025-zoomeye}, DeepEyes~\citep{zheng2025deepeyes}, and DeepEyesV2~\citep{hong2025deepeyesv2}. These workflow-based systems are particularly relevant because they also aim to improve fine-grained perception by changing how visual evidence is acquired or processed.

\paragraph{Benchmarks.}
Since fine-grained high-resolution perception tasks naturally test a model's ability to observe visual details, we evaluate models on three popular benchmarks: V*~\citep{wu2024v}, HR-Bench 4K, and HR-Bench 8K~\citep{wang2025divide}. Details of these benchmarks are provided in Appendix~\ref{app:dataset-benchmark-details}, and the evaluation protocol is described in Appendix~\ref{app:evaluation-protocol}.

\paragraph{Training Details.}

We train Qwen3-VL-Instruct~\citep{bai2025qwen3} with OPO for 200 iterations on 8 NVIDIA H20 GPUs. Each training batch contains 64 prompts, and each prompt uses a rollout budget of $M=8$, consisting of $B=2$ initial rollouts and $P=6$ remaining trajectory slots for observation-event branching and backfill.
For each sample, the model can generate up to five assistant turns and invoke the image zoom-in tool during reasoning. We follow the tool-call format of Qwen3-VL; the detailed prompt template is provided in Appendix~\ref{app:prompt-tool-configuration}. For fair comparison, we use the same training data as DeepEyes~\citep{zheng2025deepeyes}, with data details provided in Appendix~\ref{app:training-data-details}. Additional rollout and optimization details are provided in Appendix~\ref{app:implementation-details}.

\subsection{Main Results}

\begin{table}[t]
\centering
\caption{
Main results on high-resolution perception benchmarks.
LLaVA-OV abbreviates LLaVA-OneVision.
Tool denotes the inference-time mechanism: Zoom for image zoom-in crop tools, WF for manually designed visual-search workflows, Code for code-execution tools, General for code-execution, image search and text search tools, C+S for image cropping and search, $\times$ for no tool.
Best and second-best results are bolded and underlined.
$^\dagger$ denotes results reproduced by us, and -- indicates that the corresponding entry is not reported or not applicable.
}
\label{tab:main-results}
\resizebox{\linewidth}{!}{
\renewcommand{\arraystretch}{1.1}
\setlength{\tabcolsep}{3pt}
\begin{tabular}{@{}llcc|ccc|ccc|ccc@{}}
\toprule
\textbf{Model} & \textbf{Backbone} & \textbf{Tool} & \textbf{Param.} &
\multicolumn{3}{c|}{\textbf{V* Bench}} &
\multicolumn{3}{c|}{\textbf{HR-Bench 4K}} &
\multicolumn{3}{c}{\textbf{HR-Bench 8K}} \\
 & & & &
\textbf{Attr.} & \textbf{Spatial} & \textbf{Overall} &
\textbf{FSP} & \textbf{FCP} & \textbf{Overall} &
\textbf{FSP} & \textbf{FCP} & \textbf{Overall} \\
\midrule
\multicolumn{13}{l}{\textit{Advanced proprietary models}} \\
GPT-5 mini~\citep{openai2025gpt5systemcard} & -- & -- & -- & -- & -- & 78.6 & -- & -- & 78.6 & -- & -- & 74.4 \\
GPT-5.2~\citep{openai2025gpt52systemcard} & -- & -- & -- & -- & -- & 75.9 & -- & -- & -- & -- & -- & -- \\
Claude Opus 4.6~\citep{anthropic2026claudeopus46} & -- & -- & -- & -- & -- & 66.5 & -- & -- & -- & -- & -- & -- \\
Gemini-3 Pro~\citep{googledeepmind2025gemini3pro} & -- & -- & -- & -- & -- & 88.0 & -- & -- & -- & -- & -- & -- \\
\midrule
\multicolumn{13}{l}{\textit{State-of-the-art open-source models}} \\
LLaVA-OV~\citep{li2024llavaonevisioneasyvisualtask} & -- & $\times$ & 7B & 75.7 & 75.0 & 75.4 & 72.0 & 54.0 & 63.0 & 67.3 & 52.3 & 59.8 \\
Qwen2.5-VL~\citep{bai2025qwen25vltechnicalreport} & -- & $\times$ & 7B & 73.9 & 67.1 & 71.2 & 85.2 & 52.2 & 68.8 & 78.8 & 51.8 & 65.3 \\
Qwen2.5-VL~\citep{bai2025qwen25vltechnicalreport} & -- & $\times$ & 32B & 87.8 & 88.1 & 87.9 & 89.8 & 58.0 & 73.9 & 84.5 & 56.3 & 70.4 \\
Qwen3-VL~\citep{bai2025qwen3} & -- & Zoom & 235B-A22B & -- & -- & \underline{93.7} & -- & -- & \underline{85.4} & -- & -- & \textbf{82.4} \\
Qwen3-VL~\citep{bai2025qwen3} & -- & Zoom & 30B-A3B & -- & -- & 89.5 & -- & -- & 82.5 & -- & -- & 79.3 \\
Qwen3-VL~\citep{bai2025qwen3} & -- & Zoom & 32B & -- & -- & 91.1 & -- & -- & 84.6 & -- & -- & 81.6 \\
Qwen3.5~\citep{qwen35blog} & -- & -- & 397B-A17B & -- & -- & 91.1 & -- & -- & -- & -- & -- & -- \\
Qwen3.5~\citep{qwen35blog} & -- & Code & 397B-A17B & -- & -- & \textbf{95.8} & -- & -- & -- & -- & -- & -- \\
Kimi K2.5~\citep{kimiteam2026kimik25visualagentic} & -- & -- & 1T-A32B & -- & -- & 77.0 & -- & -- & -- & -- & -- & -- \\
\midrule
\multicolumn{13}{l}{\textit{Specialized visual reasoning methods}} \\
SEAL~\citep{wu2024v} & LLaVA & WF & 7B & 74.8 & 76.3 & 75.4 & -- & -- & -- & -- & -- & -- \\
DyFo~\citep{li2025dyfo} & Qwen2-VL & WF & 7B & 80.0 & 82.9 & 81.2 & -- & -- & -- & -- & -- & -- \\
ZoomEye~\citep{shen-etal-2025-zoomeye} & LLaVA-OV & WF & 7B & \textbf{93.9} & 85.5 & 90.6 & 84.3 & 55.0 & 69.6 & \underline{88.5} & 50.0 & 69.3 \\
DeepEyes~\citep{zheng2025deepeyes} & Qwen2.5-VL & Zoom & 7B & \underline{91.3} & \underline{88.2} & 90.1 & 91.3 & 59.0 & 75.1 & 86.8 & 58.5 & 72.6 \\
DeepEyesV2~\citep{hong2025deepeyesv2} & Qwen2.5-VL & General & 7B & -- & -- & 81.8 & -- & -- & 77.9 & -- & -- & 73.8 \\
Thyme-RL~\citep{zhang2025thyme} & Qwen2.5-VL & Code & 7B & 83.5 & 80.3 & 82.2 & 91.0 & 63.0 & 77.0 & 86.5 & 57.5 & 72.0 \\
ZwZ-8B~\citep{wei2026zwz} & Qwen3-VL & $\times$ & 8B & -- & -- & 91.1 & -- & -- & 84.4 & -- & -- & \underline{82.0} \\
SenseNova-MARS~\citep{chng2025sensenovamars} & Qwen3-VL & C+S & 8B & -- & -- & 92.2 & -- & -- & 83.1 & -- & -- & 78.4 \\
\midrule
Qwen3-VL-2B$^\dagger$~\citep{bai2025qwen3} & -- & Zoom & 2B & 82.6 & 65.8 & 75.9 & 79.3 & 52.8 & 66.0 & 68.0 & 53.3 & 60.6 \\
\rowcolor{blue!6}
\quad \textbf{+ \method} & Qwen3-VL & Zoom & 2B & 89.6 & 80.3 & 85.9 & 90.5 & 66.3 & 78.4 & 85.3 & \underline{67.5} & 76.4 \\
\rowcolor{blue!6}
\quad $\Delta$ (vs Qwen3-VL-2B) & -- & -- & -- & +7.0 & +14.5 & +10.0 & +11.2 & +13.5 & +12.4 & +17.3 & +14.2 & +15.8 \\
Qwen3-VL-8B$^\dagger$~\citep{bai2025qwen3} & -- & Zoom & 8B & \underline{91.3} & 82.9 & 88.0 & \underline{92.5} & \underline{70.3} & 81.4 & 87.8 & 65.3 & 76.5 \\
\rowcolor{blue!6}
\quad \textbf{+ \method} & Qwen3-VL & Zoom & 8B & \textbf{93.9} & \textbf{90.8} & 92.7 & \textbf{95.8} & \textbf{75.3} & \textbf{85.5} & \textbf{93.0} & \textbf{68.5} & 80.8 \\
\rowcolor{blue!6}
\quad $\Delta$ (vs Qwen3-VL-8B) & -- & -- & -- & +2.6 & +7.9 & +4.7 & +3.3 & +5.0 & +4.1 & +5.3 & +3.3 & +4.3 \\
\bottomrule
\end{tabular}
}
\end{table}
For readability, we denote Qwen3-VL-2B/8B trained with \method\ as OPO-2B/8B.

\paragraph{Overall performance.}
Table~\ref{tab:main-results} presents the main results of \method\ on high-resolution perception benchmarks. Overall, \method\ consistently improves performance across model scales and datasets. Based on Qwen3-VL-2B, OPO-2B improves the overall accuracy on V*, HR-Bench 4K, and HR-Bench 8K by 10.0, 12.4, and 15.8 points, respectively. Based on Qwen3-VL-8B, OPO-8B also brings gains of 4.7, 4.1, and 4.3 points. The gains suggest that observation-action optimization is particularly effective for acquiring fine-grained visual evidence in high-resolution settings. These results show that \method\ substantially improves smaller models while remaining effective at larger scales.

\paragraph{Comparison with strong baselines.}
Compared with strong open-source and proprietary models, OPO-8B also achieves competitive performance. On V*, OPO-8B obtains 92.7\% overall accuracy, outperforming several strong baselines, including Gemini-3 Pro, Qwen3-VL-32B, and Qwen3.5, while remaining below the much larger Qwen3-VL-235B-A22B and the Code-enabled Qwen3.5. On HR-Bench 4K, OPO-8B reaches 85.5\% overall accuracy, achieving the best result in the table and slightly surpassing Qwen3-VL-235B-A22B at 85.4\%. On HR-Bench 8K, OPO-8B improves Qwen3-VL-8B from 76.5\% to 80.8\%, approaching the performance of much larger models. These results suggest that directly optimizing observation actions can effectively improve high-resolution visual evidence acquisition, rather than relying solely on scaling model size.

\paragraph{Comparison with specialized visual reasoning methods.}
OPO-8B also compares favorably with other specialized visual reasoning methods. SEAL, DyFo, ZoomEye, DeepEyes, DeepEyesV2, Thyme-RL, and SenseNova-MARS improve fine-grained visual perception through visual search, dynamic focusing, or tool use, while ZwZ distills region-level supervision into the model for single-pass perception without tools. OPO-8B achieves 92.7\% and 85.5\% overall accuracy on V* and HR-Bench 4K, respectively, ranking first among these methods on both benchmarks; on HR-Bench 8K, it achieves 80.8\%, second only to ZwZ-8B. Notably, OPO-2B achieves 78.4\% and 76.4\% on HR-Bench 4K and 8K, outperforming several 7B baselines. These results show that targeted observation exploration and localized credit attribution enable OPO to improve the acquisition and use of fine-grained visual evidence.

\begin{table}[t]
\centering
\caption{
Controlled comparison between DeepEyes-style outcome-level GRPO and \method\ under the same backbone and zoom-tool training setup.
All results are reproduced by us. }
\label{tab:deepeyes-style-grpo}
\resizebox{\linewidth}{!}{
\renewcommand{\arraystretch}{1.1}
\begin{tabular}{l|ccc|ccc|ccc}
\toprule
\textbf{Model} &
\multicolumn{3}{c|}{\textbf{V* Bench}} &
\multicolumn{3}{c|}{\textbf{HR-Bench 4K}} &
\multicolumn{3}{c}{\textbf{HR-Bench 8K}} \\
 & \textbf{Attr.} & \textbf{Spatial} & \textbf{Overall}
 & \textbf{FSP} & \textbf{FCP} & \textbf{Overall}
 & \textbf{FSP} & \textbf{FCP} & \textbf{Overall} \\
\midrule
Qwen2.5-VL-7B~\citep{bai2025qwen25vltechnicalreport} & 53.0 & 55.3 & 53.9 & 68.5 & 55.3 & 61.9 & 60.8 & 48.0 & 54.4 \\
\quad + DeepEyes-style GRPO~\citep{zheng2025deepeyes} & 87.0 & 81.6 & 84.8 & 88.0 & 55.0 & 71.5 & 76.3 & 59.8 & 68.0 \\
\rowcolor{blue!6}
\quad + \method & 93.9 & 80.3 & 88.5 & 87.8 & 60.3 & 74.0 & 78.8 & 57.5 & 68.1 \\
\rowcolor{blue!6}
\quad $\Delta$ (vs GRPO)& +6.9 & -1.3 & +3.7 & -0.2 & +5.3 & +2.5 & +2.5 & -2.3 & +0.1 \\
\midrule
Qwen3-VL-2B~\citep{bai2025qwen3} & 82.6 & 65.8 & 75.9 & 79.3 & 52.8 & 66.0 & 68.0 & 53.3 & 60.6 \\
\quad + DeepEyes-style GRPO~\citep{zheng2025deepeyes} & 85.2 & 79.0 & 82.7 & 88.0 & 55.0 & 71.5 & 85.3 & 57.0 & 71.1 \\
\rowcolor{blue!6}
\quad + \method & 89.6 & 80.3 & 85.9 & 90.5 & 66.3 & 78.4 & 85.3 & 67.5 & 76.4 \\
\rowcolor{blue!6}
\quad $\Delta$ (vs GRPO) & +4.4 & +1.3 & +3.2 & +2.5 & +11.3 & +6.9 & +0.0 & +10.5 & +5.3 \\
\midrule
Qwen3-VL-8B~\citep{bai2025qwen3} & 91.3 & 82.9 & 88.0 & 92.5 & 70.3 & 81.4 & 87.8 & 65.3 & 76.5 \\
\quad + DeepEyes-style GRPO~\citep{zheng2025deepeyes} & 87.8 & 93.4 & 90.1 & 91.5 & 73.3 & 82.4 & 86.3 & 69.3 & 77.8 \\
\rowcolor{blue!6}
\quad + \method & 93.9 & 90.8 & 92.7 & 95.8 & 75.3 & 85.5 & 93.0 & 68.5 & 80.8 \\
\rowcolor{blue!6}
\quad $\Delta$ (vs GRPO) & +6.1 & -2.6 & +2.6 & +4.3 & +2.0 & +3.1 & +6.7 & -0.8 & +3.0 \\
\bottomrule
\end{tabular}
}
\end{table}

\paragraph{Comparison with outcome-level GRPO.}
Table~\ref{tab:deepeyes-style-grpo} further provides a controlled comparison between \method\ and DeepEyes-style outcome-level GRPO. The DeepEyes results in Table~\ref{tab:main-results} are taken from the original paper, whereas the DeepEyes-style GRPO baselines in Table~\ref{tab:deepeyes-style-grpo} are retrained by us under the same backbone, training budget, and zoom-tool setup as OPO. DeepEyes-style GRPO relies only on outcome-level rewards such as final-answer correctness, format quality, and tool-use signals. In contrast, \method\ further introduces uncertainty-guided local branching and evidence-localized attribution. 
Although DeepEyes-style GRPO performs better on a few subtasks, \method\ is more stable in terms of overall accuracy, especially on HR-Bench. This suggests that outcome-level rewards can encourage the model to use the zoom-in tool, but they still struggle to assign local evidence-quality feedback to the specific observation actions. By exploring alternative crops around high-uncertainty observation events and applying evidence-sufficiency feedback to the corresponding observation-action tokens, OPO more effectively optimizes where the model looks.

Qualitative examples and analysis are provided in Appendix~\ref{app:qualitative-results} and Figures~\ref{fig:qualitative-results-1}--\ref{fig:qualitative-results-2}.

\subsection{Ablation Study}

\begin{wraptable}{R}{0.48\linewidth}
\centering
\scriptsize
\captionsetup{skip=4pt}

\caption{
Ablation study on Qwen3-VL-2B. 
We report overall accuracy. 
The default setting is $M=8,B=2$, where $M$ is the rollout budget and $B$ is the number of base trajectories. 
The default full OPO setting is highlighted in bold.
}
\label{tab:ablation}
\setlength{\tabcolsep}{3pt}
\renewcommand{\arraystretch}{1.1}

\begin{tabular*}{\linewidth}{@{\extracolsep{\fill}}lccc@{}}
\toprule
\textbf{Variant} & \textbf{V*} & \textbf{HR-Bench 4K} & \textbf{HR-Bench 8K} \\
\midrule
Qwen3-VL-2B & 75.9 & 66.0 & 60.6 \\
OPO full & \textbf{85.9} & \textbf{78.4} & \textbf{76.4} \\
\midrule
w/o UOB \& ELAA & 82.7 & 71.5 & 71.1 \\
w/o ELAA & 83.8 & 76.5 & 73.8 \\
w/o uncert. sel. & 83.3 & 76.9 & 73.0 \\
w/o redund. filt. & 84.8 & 76.5 & 75.1 \\
\midrule
$M=4,B=1$  & 82.4 & 74.2 & 72.5 \\
$M=8,B=1$  & 84.3 & 76.6 & 74.8 \\
$M=8,B=2$  & \textbf{85.9} & \textbf{78.4} & \textbf{76.4} \\
$M=8,B=4$  & 84.8 & 77.1 & 75.3 \\
$M=16,B=2$ & 86.2 & 78.8 & 76.9 \\
\bottomrule
\end{tabular*}
\par\vspace{-3.5pt}
\end{wraptable}

\paragraph{Effect of core modules.}
Table~\ref{tab:ablation} presents the ablation results based on Qwen3-VL-2B. Unless otherwise specified, all \method\ variants use the default rollout setting $M=8,B=2$, where $M$ is the total rollout budget, $B$ is the number of base trajectories, and the remaining $P=M-B$ slots are used for event-local branching and backfill. Under the default rollout budget, the full \method\ achieves the best performance among all module ablations, showing that both uncertainty-guided observation exploration and evidence-localized attribution are important for improving high-resolution visual evidence acquisition.

Since ELAA relies on sibling trajectories produced by UOB for local comparison, we do not remove UOB alone; instead, we jointly remove UOB and ELAA. Without UOB and ELAA, the model degenerates into DeepEyes-style outcome-level GRPO training~\citep{zheng2025deepeyes}. This variant performs substantially worse than the full \method\ on all three benchmarks, suggesting that ordinary full-trajectory sampling is insufficient for exploring alternative observation actions, especially in high-resolution tasks where an incorrect crop can directly affect the final answer.

Removing ELAA keeps the UOB branching mechanism but removes localized evidence attribution, so the evidence-sufficiency feedback is no longer applied specifically to observation-action tokens and instead becomes a trajectory-level signal. This variant still shows consistent degradation compared with the full method, indicating that outcome-level or trajectory-level supervision alone can cause credit mismatch between observation actions and downstream reasoning. Meanwhile, w/o ELAA remains better than w/o UOB \& ELAA, suggesting that UOB's local branching itself is beneficial, but ELAA is needed to more effectively convert local evidence quality into observation-action learning signals.

We further examine the design choices inside UOB. Replacing uncertainty-guided event selection with random branching reduces performance, indicating that high-entropy observation events are more informative branching points than randomly selected zoom events. Removing redundancy filtering also hurts performance, suggesting that without penalizing crops that heavily overlap with previous observations, the limited branching budget can be wasted on redundant visual regions.

\paragraph{Effect of rollout allocation.}
We also analyze the effect of rollout budget and base-trajectory allocation. The smaller setting $(M=4,B=1)$ performs worse, showing that insufficient rollout budget limits local observation exploration. When fixing $M=8$, using only $B=1$ provides too few base candidates and may miss useful observation events, while increasing to $B=4$ reduces the budget available for local branching and limits the exploration of alternative crops. The default setting $M=8,B=2$ achieves a better balance between base-candidate coverage and event-local branching. Increasing the budget to $M=16,B=2$ brings only marginal gains, suggesting diminishing returns from larger rollout budgets.

\section{Conclusion}

In this work, we study a key challenge in tool-augmented thinking with images: models must learn not only how to reason, but also where to look during reasoning. We show that many failures in high-resolution fine-grained visual tasks originate from the observation stage rather than downstream language reasoning, and that the observation-action tokens determining local crops exhibit substantially higher uncertainty than ordinary language tokens. These findings suggest that visual evidence acquisition is a critical decision-making process that should be optimized directly, rather than treated as an ordinary part of trajectory-level generation.
Based on these observations, we propose Observation Policy Optimization (OPO), which improves visual evidence acquisition through uncertainty-guided observation branching and evidence-localized advantage attribution. OPO performs local branching at high-uncertainty and low-redundancy observation events, allowing the model to explore alternative crops from shared reasoning prefixes. It then assigns evidence-sufficiency feedback only to the observation-action tokens that produce the corresponding crops, thereby reducing the credit mismatch between final-answer rewards and local evidence quality.
Experiments across high-resolution visual reasoning benchmarks show that OPO consistently improves Qwen3-VL-2B and Qwen3-VL-8B, and OPO-8B achieves competitive performance against much larger models and tool-augmented baselines. These results highlight observation policy optimization as an important direction for improving active visual reasoning.

\clearpage
\begin{ack}
This work was supported by the Ministry of Industry and Information Technology under Grant No. BB2320000141; the National Key Research and Development Plan under Grant No. 2024YFC3014700; the Central Government-Guided Local Science and Technology Development Funds of Anhui Province under Grant No. 2023CSJGG1100; the National Natural Science Foundation of China (NSFC) under Grant No. 62476260; the Fundamental Research Funds for the Central Universities under Grant No. WK2100000057; and the CCF-1688 Yuanbao Cooperation Fund under Grant No. 2025006. The numerical computations were performed on the supercomputing system at the Supercomputing Center of the University of Science and Technology of China.
\end{ack}

\bibliographystyle{unsrtnat}
\bibliography{my_nips}

\appendix

\clearpage

\section{Algorithm}

\subsection{Algorithm Procedure}
\label{app:algorithm-procedure}

Algorithm~\ref{alg:opo} summarizes the full \method\ training procedure, combining the rollout-time observation branching of UOB (Section~\ref{sec:uob}) with the evidence-localized advantage assignment of ELAA (Section~\ref{sec:elaa}).

\begin{algorithm}[ht]
\caption{\method\ with UOB and ELAA}
\label{alg:opo}
\begin{algorithmic}[1]
\Require Policy $\pi_\theta$, input sample $x$, total rollout budget $M$, base rollout budget $B$, branch budget $P=M-B$
\Statex \textbf{UOB: rollout-time observation branching}
\State Generate $B$ base rollouts until the first valid image zoom-in call or termination.
\State For each candidate observation event $e$, record its pre-tool prefix, observation-action spans, and the original candidate crop.
\State Estimate bounding-box uncertainty $g(e)$ from the bounding-box token span.
\State Compute the question-normalized anomaly score $z(e)$.
\For{each candidate event e while the branch budget remains do}
    \State Allocate at most $Z(e)=\mathcal{A}(z(e),P_{\mathrm{rem}})$ child proposals.
    \State Initialize $\mathcal{C}_e$ with the original candidate crop.
    \State Preserve the prompt and pre-tool prefix, then resample $Z(e)$ suffixes from that state.
    \State Sequentially accept valid child candidates with $\mathrm{Red}(c\mid e)\leq\tau_{\mathrm{red}}$ and add them to $\mathcal{C}_e$.
    \State Execute accepted child tool calls, continue each retained trajectory to termination, and form a branch group $g$ from trajectories sharing the same pre-tool prefix.
\EndFor
\State Backfill with ordinary rollouts until $M$ complete trajectories are obtained.
\Statex \textbf{ELAA: evidence-localized advantage attribution}
\State Compute global outcome rewards $R_i^{\mathrm{global}}$ and standard GRPO advantages $A_i^{\mathrm{global}}$ by question id.
\State For each branch group $g$, judge each crop $c_{i,g}$ and compute group-local evidence rewards $R_{i,g}^{\mathrm{evi}}=\rho(c_{i,g})$.
\State Normalize $R_{i,g}^{\mathrm{evi}}$ within $\mathcal{S}(g)$ to obtain $A_{i,g}^{\mathrm{local}}$.
\State Apply $A_{i,g}^{\mathrm{local}}$ only to image-index, region-label, and bounding-box spans; use $A_i^{\mathrm{global}}$ for other generated tokens.
\State Update $\pi_\theta$ with the clipped PPO policy objective.
\end{algorithmic}
\end{algorithm}

\subsection{Policy Update}
\label{app:policy-update}

After obtaining the token-level advantage $A_{i,t}$, \method\ updates the policy with the standard clipped PPO objective. OPO does not change the form of the PPO update; its effect enters through the advantage $A_{i,t}$ defined in the main text, which assigns global outcome advantages to ordinary tokens and evidence-local advantages to observation-action tokens.

For generated token $y_{i,t}$, let
\begin{equation}
 r_{i,t}(\theta)=
 \frac{\pi_\theta(y_{i,t}\mid x_i,y_{i,<t})}
 {\pi_{\theta_{\mathrm{old}}}(y_{i,t}\mid x_i,y_{i,<t})}.
\end{equation}
The policy objective is
\begin{equation}
\mathcal{L}_{\mathrm{policy}}
= -\sum_{i,t} m_{i,t}^{\mathrm{resp}}
\min\left(
 r_{i,t}(\theta)A_{i,t},
 \mathrm{clip}(r_{i,t}(\theta),1-\epsilon_{\mathrm{clip}},1+\epsilon_{\mathrm{clip}})A_{i,t}
\right),
\end{equation}
where $m_{i,t}^{\mathrm{resp}}$ is the response mask. Tool-observation tokens returned by the environment are masked out and do not contribute to the policy loss.

\section{Benchmark Details}
\label{app:dataset-benchmark-details}

We evaluate \method\ on three high-resolution perception benchmarks:
V*~\citep{wu2024v}, HR-Bench 4K, and HR-Bench 8K~\citep{wang2025divide}.
Following the high-resolution evaluation setting used in DeepEyes~\citep{zheng2025deepeyes}, these benchmarks focus on the ability of models to acquire and interpret fine-grained visual evidence from high-resolution images. Many examples cannot be answered reliably from a global image view or a heavily downsampled image alone; instead, they require models to locate local evidence such as small objects, subtle attributes, local text, or spatial relations.

V*~\citep{wu2024v} focuses on guided visual search over natural images and evaluates whether a model can locate and use critical visual details in high-resolution images. The V* split used in our evaluation contains 191 examples in total, including 115 examples from \texttt{direct\_attributes} and 76 examples from \texttt{relative\_position}. For V*, we report the benchmark-defined attribute, spatial, and overall accuracies.

HR-Bench~\citep{wang2025divide} contains two high-resolution evaluation versions, HR-Bench 4K and HR-Bench 8K, which measure whether models can resolve local visual evidence under increasingly demanding image resolutions. Each version contains 200 original questions, including 100 Fine-grained Single-instance Perception (FSP) questions and 100 Fine-grained Cross-instance Perception (FCP) questions. FSP evaluates fine-grained perception over a single instance, such as attribute recognition, OCR, and visual prompting, while FCP evaluates fine-grained perception across multiple instances, such as map analysis, chart analysis, and spatial relationship assessment.

In our evaluation, each original HR-Bench question is expanded into four option-rotation instances. As a result, \texttt{hr\_bench\_4k} contains 800 evaluation rows and \texttt{hr\_bench\_8k} contains 800 evaluation rows, yielding 1,600 rows in total. When reporting results, we aggregate performance according to the original benchmark definitions and report FSP, FCP, and overall accuracy following the official scoring protocol.

These benchmarks are particularly aligned with our setting because successful prediction often requires an explicit observation action: the model must first decide which local region should be inspected and then answer based on the acquired local evidence. Therefore, for both OPO and our reproduced Qwen3-VL tool-use baselines, we keep the image zoom-in tool enabled during inference. The model is allowed to perform multi-turn tool-augmented reasoning, and we evaluate only the final answer after the trajectory terminates.

\section{Evaluation Protocol}
\label{app:evaluation-protocol}

We evaluate saved checkpoints with deterministic multi-turn generation. For each benchmark example, we use the same system instructions, response format, and tool schema as in training, and generate one assistant turn at a time using greedy decoding. Each sample allows up to 10 tool-use turns. When the generated response contains a valid \texttt{<tool\_call>...</tool\_call>} block, the evaluator parses the JSON arguments, converts the bounding box to image coordinates, crops the selected image from the visible image pool, appends the crop as a tool message, and continues generation. The trajectory terminates when an \texttt{<answer>...</answer>} block is produced, an invalid tool call is encountered, or the maximum number of turns is reached.

For V*, we evaluate the direct attributes and relative position splits. Each example is treated as a multiple-choice visual question, with the first option as the benchmark-provided correct answer. The final answer is extracted from the answer tags. We apply an LLM judge to compare the model answer with the standard answer using few-shot consistency examples and return a binary judgement. We report accuracy for each split and compute the V* overall score as the mean over all evaluated examples.

For HR-Bench, we evaluate the HR-Bench 4K and HR-Bench 8K splits. Images are decoded from the benchmark TSV files and resized according to the Qwen-VL visual grid before inference. The same multi-turn tool loop and answer extraction procedure are used. The judging script reports accuracy separately for the benchmark's Fine-grained Single-instance Perception (FSP) and Fine-grained Cross-instance Perception (FCP) categories and computes each split's overall score as the unweighted average of these two category accuracies. These scores correspond to the HR-Bench columns reported in Table~\ref{tab:main-results}.

\section{Training Data Details}
\label{app:training-data-details}

For fair comparison with DeepEyes, we use the same tool-oriented training data recipe rather than constructing a new training corpus. As described in DeepEyes~\citep{zheng2025deepeyes}, the training mixture contains about 47K samples from three complementary sources: Visual Search, ArxivQA, and ThinkLite-VL. Visual Search contributes about 22K samples, roughly 47\% of the mixture, and emphasizes natural-image grounding and fine-grained perception. ArxivQA contributes about 14K samples, roughly 30\%, covering scientific plots, diagrams, and chart-style visual inputs. ThinkLite-VL contributes about 11K samples, roughly 23\%, and provides multimodal reasoning examples involving arithmetic, commonsense inference, and problem solving.

The DeepEyes data curation process follows a tool-use-oriented filtering pipeline. First, it removes questions that are either too easy or too difficult by estimating model accuracy from multiple responses. Second, it converts questions into an open-ended format and filters unreliable conversions. Third, it removes unverifiable samples, including examples with incorrect answers or unreadable questions. Finally, it prioritizes samples where tool use can improve visual evidence acquisition, especially cases where the model fails from the original image but can answer correctly when given the relevant cropped region. Chart data and reasoning data are retained according to their own preprocessing assumptions because they already provide complementary structured-visual and reasoning supervision.

\section{Prompt and Tool Configuration}
\label{app:prompt-tool-configuration}

We use the Qwen3-VL tool-calling format in both training and evaluation. Each example is converted into a conversation containing tool instructions and the user's visual question. The tool instructions provide the \texttt{image\_zoom\_in\_tool} function signature inside \texttt{<tools>...</tools>} XML tags and require each function call to be returned as a JSON object inside \texttt{<tool\_call>...</tool\_call>} tags. The user message contains the image placeholder, the original question, candidate options, and the instruction to think first, call tools when needed, and finally provide an answer enclosed by answer tags. The concrete prompt and tool-call schema are summarized in Figure~\ref{fig:prompt_tool_schema}.

\begin{figure}[!t]
\centering
\captionsetup{skip=4pt}
\begin{minipage}{0.95\linewidth}
\begin{promptbox}
SYSTEM:
You are a helpful assistant. You can call functions to assist with the user query.
Important: You must call only one function at a time. After each function call,
wait for the execution result before making the next function call if needed.

# Tools

You may call one or more functions to assist with the user query.

You are provided with function signatures within <tools></tools> XML tags:
<tools>
{
  "type": "function",
  "function": {
    "name": "image_zoom_in_tool",
    "description": "Zoom in on a specific region of an image by cropping it based on a bounding box (bbox) and an optional object label.",
    "parameters": {
      "type": "object",
      "properties": {
        "bbox_2d": {
          "type": "array",
          "items": {"type": "number"},
          "minItems": 4,
          "maxItems": 4,
          "description": "The bounding box of the region to zoom in, as [x1, y1, x2, y2], where (x1, y1) is the top-left corner and (x2, y2) is the bottom-right corner."
        },
        "label": {
          "type": "string",
          "description": "The name or label of the object in the specified bounding box."
        },
        "img_idx": {
          "type": "integer",
          "description": "0-based index into the currently visible image list, including the original input images followed by tool-returned crops from earlier turns."
        }
      },
      "required": ["bbox_2d", "label", "img_idx"]
    }
  }
}
</tools>

For each function call, return a json object with function name and arguments within <tool_call></tool_call> XML tags:
<tool_call>
{"name": <function-name>, "arguments": <args-json-object>}
</tool_call>

USER:
<image>
Your role is that of a research assistant specializing in visual information. Answer questions about images by looking at them closely and then using research tools. Please follow this structured thinking process and show your work.

Start an iterative loop for each question:
- First, look closely: Begin with a detailed description of the image, paying attention to the user's question. List what you can tell just by looking, and what you'll need to look up.
- Next, find information: Use a tool to research the things you need to find out.
- Then, review the findings: Carefully analyze what the tool tells you and decide on your next action.

Continue this loop until your research is complete.

To finish, bring everything together in a clear, synthesized answer that fully responds to the user's question.

Question: {question}
Options: 
A. {option_1}
B. {option_2}
...

First think in <thinking>...</thinking>. If needed, call tools using
<tool_call>...</tool_call>. Final answer must be enclosed in <answer>...</answer>.
\end{promptbox}
\end{minipage}
\caption{Prompt and tool-call schema used for tool-augmented visual reasoning, including the tool signature for \texttt{image\_zoom\_in\_tool} and the required \texttt{<tool\_call>} return format.}
\label{fig:prompt_tool_schema}
\end{figure}

The image zoom-in tool crops a region from the selected image and returns the cropped image as the next tool observation. The tool maintains a visible image pool consisting of the original input image followed by all tool-returned crops from previous turns. Thus, \texttt{img\_idx=0} refers to the original image, while later indices can refer to earlier crops. For Qwen3-VL models, bounding boxes are interpreted in the normalized $[0,1000]$ coordinate system and converted to image-space pixels before cropping.

\section{Reward and Evidence Judge}
\label{app:reward-evidence-judge}

The reward function has two parts: a trajectory-level final-answer reward and an event-local evidence reward. The final-answer reward first extracts the model's answer from \texttt{<answer>...</answer>} after the thinking segment. A judge then compares the extracted answer with the ground-truth answer given the original question and returns a binary correctness label. We denote this answer reward by $r_{\mathrm{acc}}\in\{0,1\}$. The implementation also includes a tool-use reward $r_{\mathrm{tool}}$, which is one only when the trajectory uses the zoom-in tool and the final answer is correct, and a format reward $r_{\mathrm{fmt}}$, which is $-1$ when the response violates the required thinking/answer format and $0$ otherwise. The global trajectory reward is
\begin{equation}
R^{\mathrm{global}} = 0.8 r_{\mathrm{acc}} + 1.2 r_{\mathrm{tool}} + 0.2 r_{\mathrm{fmt}}.
\end{equation}
This global reward is used for the ordinary trajectory-level GRPO signal.

For OPO's local observation-action supervision, we additionally use an evidence judge. For each branch event, the training pipeline keeps the crops associated with that event and sends each crop, together with the question and ground-truth answer, to a frozen visual judge. The judge must return exactly one of three labels: \texttt{insufficient}, \texttt{partial}, or \texttt{sufficient}. These labels are mapped to local evidence rewards $-1$, $0$, and $1$, respectively. 

The exact prompts for the final-answer judge and the local evidence judge are shown in Figure~\ref{fig:reward_judge_prompts}. In our implementation, the final-answer reward judge is instantiated with Qwen3-Plus, while the local evidence reward judge is instantiated with Qwen3-VL-Max.

\begin{figure}[!t]
\centering
\begin{minipage}{0.95\linewidth}
\begin{promptbox}
LLM ANSWER JUDGE:
You are an expert evaluator. Your task is to determine if a model's answer is
semantically equivalent to a provided standard answer, given a specific question.
Your evaluation must be strict. The model's answer is only correct if it fully
matches the meaning of the standard answer.
You must provide your final judgement as a single word: either "CORRECT" or
"INCORRECT". Do not provide any explanation or other text.

VLM EVIDENCE JUDGE:
You are a visual evidence judge.

Your task is NOT to answer the question.
Your task is to decide whether the current cropped image contains enough visible
evidence to support the given ground-truth answer.

Inputs:
1. Question
2. Ground-truth answer
3. Current cropped image

Judging rule:
Decide whether the cropped image provides sufficient visual evidence for the
ground-truth answer.

Labels:
- insufficient: the crop clearly does not contain enough visual evidence to
  support the answer
- partial: the crop contains some relevant evidence, but it is still incomplete,
  ambiguous, or lacks enough context/detail to confidently support the answer
- sufficient: the crop contains enough clear visual evidence to support the answer

Important constraints:
- Judge only from the cropped image.
- Do NOT rely on commonsense, outside knowledge, or guess from the answer itself.
- Even if the ground-truth answer looks plausible, if the crop does not visibly
  support it, do NOT output sufficient.
- For spatial or relation questions, check whether enough context is preserved.
- For fine-grained attribute or OCR questions, check whether the key detail is
  visible and clear enough.

Output exactly one label from:
insufficient
partial
sufficient
\end{promptbox}
\end{minipage}
\caption{Prompts for final-answer judging and local visual-evidence judging. The LLM judge produces the trajectory-level correctness signal, while the VLM judge produces the event-local evidence reward for ELAA.}
\label{fig:reward_judge_prompts}
\end{figure}

\section{Qualitative Results}
\label{app:qualitative-results}

\begin{figure}[!t]
\centering
\includegraphics[width=\linewidth]{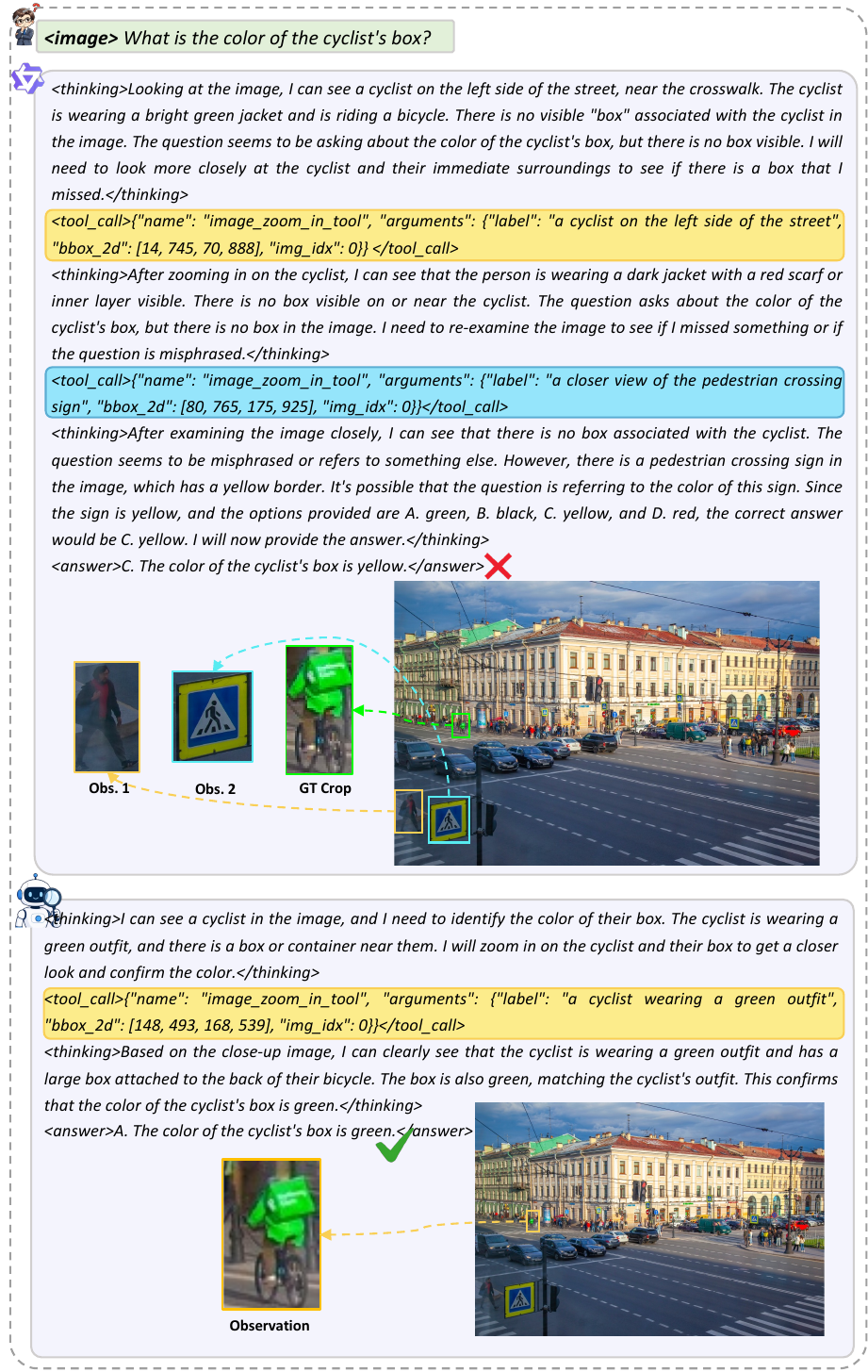}
\caption{Qualitative comparison on a V* direct-attribute example.}
\label{fig:qualitative-results-1}
\end{figure}

\begin{figure}[!t]
\centering
\includegraphics[width=\linewidth,height=0.9\textheight,keepaspectratio]{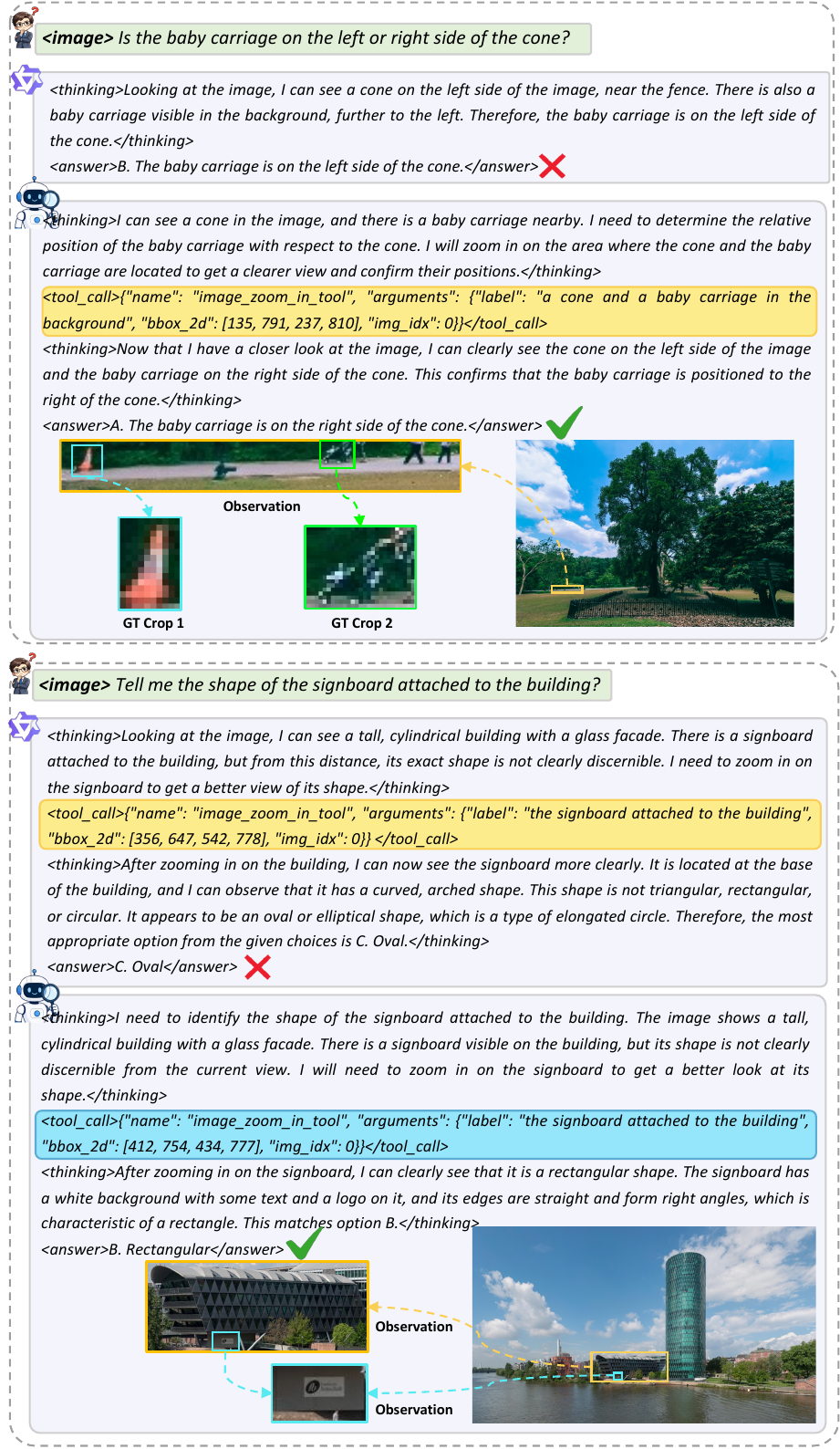}
\caption{Qualitative comparisons on V* and HR-Bench. The top example is a V* spatial-relation question, while the bottom example is from HR-Bench.}
\label{fig:qualitative-results-2}
\end{figure}

Figure~\ref{fig:qualitative-results-1} and Figure~\ref{fig:qualitative-results-2} show qualitative comparisons between direct Qwen3-VL-8B inference and OPO-8B. In the direct-attribute example (Figure~\ref{fig:qualitative-results-1}), direct Qwen3-VL-8B inference is distracted by salient but irrelevant yellow regions, such as the pedestrian-crossing sign, and mistakenly treats them as evidence for the color of the cyclist's box. In contrast, OPO-8B focuses its observation action on the cyclist and the attached box, correctly identifying the box as green from localized evidence. In the spatial-relation example (Figure~\ref{fig:qualitative-results-2}, top), Qwen3-VL-8B observes both the cone and the baby carriage but fails to reliably resolve their left-right relation; OPO-8B instead crops a local region containing both objects and correctly determines that the carriage is to the right of the cone. In the HR-Bench example (Figure~\ref{fig:qualitative-results-2}, bottom), Qwen3-VL-8B relies on a coarse building-level observation and mistakes the signboard for an oval shape, whereas OPO-8B accurately localizes the small signboard and zooms into it, allowing it to recognize its rectangular shape from the clear edges and background. These cases show that OPO's gains do not merely come from invoking the zoom-in tool, but from more reliably selecting question-relevant and evidence-sufficient observation regions, thereby grounding the final answer in localized visual evidence.

\section{Implementation Details}
\label{app:implementation-details}

In the main implementation, the rollout budget is $M=8$, with $B=2$ base rollouts and $P=6$ remaining trajectories for event-local branching and backfill. The pre-action baseline window is $W=16$, and event uncertainty is computed only from the bounding-box token span, without using image-index or post-crop tokens. Branches preserve the pre-event prefix and resample the suffix from the prompt-plus-prefix state. If the total number of trajectories remains below $M$, ordinary rollouts are used as backfill. For local attribution, the local token spans cover image-index, region-label, and bounding-box tokens, and the local advantage is computed only from the event-local evidence reward.

\section{Localization Failures in Stronger Models}
\label{app:stronger-model-localization}

To examine whether the observation bottleneck persists in stronger models, we evaluate Qwen3-VL-32B-Instruct, Qwen3.7-Max, and Kimi K3 on all 191 examples in V*. We analyze their inference trajectories without applying OPO training. We manually inspect each unsuccessful trajectory, including incorrect responses and cases without a valid final answer. A trajectory is classified as a localization failure when failure to acquire the visual evidence needed to answer the question is the primary cause of the unsuccessful response.

\begin{table}[htbp]
\centering
\small
\setlength{\tabcolsep}{4pt}
\caption{Localization failures in stronger models on V*. Percentages in the last column are computed over unsuccessful trajectories.}
\label{tab:stronger-model-localization}
\begin{tabular}{lccc}
\toprule
Model & Accuracy & Unsuccessful & Localization failures \\
\midrule
Qwen3-VL-32B-Instruct & 169/191 (88.5\%) & 22 & 17/22 (77.3\%) \\
Kimi K3 & 172/191 (90.1\%) & 19 & 15/19 (78.9\%) \\
Qwen3.7-Max & 182/191 (95.3\%) & 9 & 9/9 (100.0\%) \\
\midrule
Overall & 523/573 (91.3\%) & 50 & 41/50 (82.0\%) \\
\bottomrule
\end{tabular}
\end{table}

As shown in Table~\ref{tab:stronger-model-localization}, the three models achieve an aggregate accuracy of 91.3\%, yet localization failures account for 41 of their 50 unsuccessful trajectories (82.0\%). This pattern appears in all three models: localization failures account for 17/22 unsuccessful trajectories for Qwen3-VL-32B-Instruct, 15/19 for Kimi K3, and 9/9 for Qwen3.7-Max. Under the evaluated observation interface, acquiring fine-grained visual evidence thus remains a major source of failure even for stronger models. These findings motivate extending observation-policy optimization to larger backbones and additional model families, with controlled OPO training as a direction for future work.

\section{Limitations and Future Work}

Our work focuses on optimizing observation actions under an image zoom-in tool. This setting directly targets a key bottleneck in high-resolution fine-grained perception, where the model must decide where to look in order to acquire useful local evidence. However, tool-augmented visual reasoning in more general scenarios may involve a broader set of tools, such as OCR, detection, segmentation, search, code execution, or more complex image manipulation tools. Therefore, our current experiments do not fully characterize how OPO generalizes to a more diverse tool ecosystem.

Second, due to computational and time constraints, we train and evaluate OPO mainly on Qwen3-VL-2B and Qwen3-VL-8B, rather than further applying it to newer or stronger native multimodal backbones such as Qwen3.5. As a result, although our experiments show that OPO remains effective when scaling from 2B to 8B models, its behavior on larger and more capable backbone models remains to be further validated.

Future work can extend OPO in two main directions. First, OPO can be generalized to broader visual and multimodal tools, enabling models to learn not only where to look, but also when to invoke different tools and how to combine them for evidence acquisition. Second, applying OPO to stronger native multimodal models would further test its scalability and clarify whether observation-action optimization can complement stronger visual encoders, longer-context reasoning, and more general tool-use capabilities.

\section{Societal Impacts}

OPO aims to improve the ability of vision-language models to acquire fine-grained visual evidence from high-resolution images. This can have positive impacts in applications that require careful inspection of visual details, such as document understanding, chart analysis, scientific image interpretation, accessibility tools, and assistance with complex visual question answering. By making observation actions more reliable, the method may also reduce failures where a model answers from insufficient or irrelevant visual evidence.

At the same time, stronger tool-augmented visual reasoning systems may introduce risks if deployed without appropriate oversight. More reliable zooming and local evidence acquisition could be misused for privacy-invasive visual analysis, automated surveillance, or high-stakes decision making where model errors are difficult for users to detect. The method may also increase user trust in generated answers even when the acquired evidence is incomplete or the evidence judge is miscalibrated.

\clearpage

\clearpage
\section*{NeurIPS Paper Checklist}

\begin{enumerate}

\item {\bf Claims}
    \item[] Question: Do the main claims made in the abstract and introduction accurately reflect the paper's contributions and scope?
    \item[] Answer: \answerYes{} %
    \item[] Justification: The abstract and introduction make claims that accurately reflect the paper’s contributions and scope.

\item {\bf Limitations}
    \item[] Question: Does the paper discuss the limitations of the work performed by the authors?
    \item[] Answer: \answerYes{} %
    \item[] Justification: This paper discusses the limitations of the work in the Appendix.

\item {\bf Theory assumptions and proofs}
    \item[] Question: For each theoretical result, does the paper provide the full set of assumptions and a complete (and correct) proof?
    \item[] Answer: \answerNA{} %
    \item[] Justification: There is no theoretical results.

    \item {\bf Experimental result reproducibility}
    \item[] Question: Does the paper fully disclose all the information needed to reproduce the main experimental results of the paper to the extent that it affects the main claims and/or conclusions of the paper (regardless of whether the code and data are provided or not)?
    \item[] Answer: \answerYes{} %
    \item[] Justification: This work provides all the information needed to reproduce the main experimental results of the paper in the Implementation Details.

\item {\bf Open access to data and code}
    \item[] Question: Does the paper provide open access to the data and code, with sufficient instructions to faithfully reproduce the main experimental results, as described in supplemental material?
    \item[] Answer: \answerYes{} %
    \item[] Justification: The code is publicly available at \url{https://github.com/Yann0122/Observation-Policy-Optimization}.

\item {\bf Experimental setting/details}
    \item[] Question: Does the paper specify all the training and test details (e.g., data splits, hyperparameters, how they were chosen, type of optimizer) necessary to understand the results?
    \item[] Answer: \answerYes{} %
    \item[] Justification: The experimental setups, including data splits, training, and testing details, are provided in the Method and Experiments sections.

\item {\bf Experiment statistical significance}
    \item[] Question: Does the paper report error bars suitably and correctly defined or other appropriate information about the statistical significance of the experiments?
    \item[] Answer: \answerNo{} %
    \item[] Justification: We follow the default evaluations in the vision-language models field, which do not require error bars.

\item {\bf Experiments compute resources}
    \item[] Question: For each experiment, does the paper provide sufficient information on the computer resources (type of compute workers, memory, time of execution) needed to reproduce the experiments?
    \item[] Answer: \answerYes{} %
    \item[] Justification: We provide them in the implementation details of the main paper and the Appendix.
    
\item {\bf Code of ethics}
    \item[] Question: Does the research conducted in the paper conform, in every respect, with the NeurIPS Code of Ethics \url{https://neurips.cc/public/EthicsGuidelines}?
    \item[] Answer: \answerYes{} %
    \item[] Justification: This work conforms to the NeurIPS Code of Ethics.

\item {\bf Broader impacts}
    \item[] Question: Does the paper discuss both potential positive societal impacts and negative societal impacts of the work performed?
    \item[] Answer: \answerYes{} %
    \item[] Justification: The broader impacts are provided in the Appendix.
    
\item {\bf Safeguards}
    \item[] Question: Does the paper describe safeguards that have been put in place for responsible release of data or models that have a high risk for misuse (e.g., pre-trained language models, image generators, or scraped datasets)?
    \item[] Answer: \answerNA{} %
    \item[] Justification: The proposed method uses pre-trained models. This proposed method is safe under the safeguards of adopted pre-trained models.

\item {\bf Licenses for existing assets}
    \item[] Question: Are the creators or original owners of assets (e.g., code, data, models), used in the paper, properly credited and are the license and terms of use explicitly mentioned and properly respected?
    \item[] Answer: \answerYes{} %
    \item[] Justification: We cited the original paper that produced the code package or dataset.

\item {\bf New assets}
    \item[] Question: Are new assets introduced in the paper well documented and is the documentation provided alongside the assets?
    \item[] Answer: \answerYes{} %
    \item[] Justification: The code and accompanying documentation have been publicly released at \url{https://github.com/Yann0122/Observation-Policy-Optimization}.

\item {\bf Crowdsourcing and research with human subjects}
    \item[] Question: For crowdsourcing experiments and research with human subjects, does the paper include the full text of instructions given to participants and screenshots, if applicable, as well as details about compensation (if any)? 
    \item[] Answer: \answerNA{} %
    \item[] Justification: The paper does not involve crowdsourcing nor research with human subjects.

\item {\bf Institutional review board (IRB) approvals or equivalent for research with human subjects}
    \item[] Question: Does the paper describe potential risks incurred by study participants, whether such risks were disclosed to the subjects, and whether Institutional Review Board (IRB) approvals (or an equivalent approval/review based on the requirements of your country or institution) were obtained?
    \item[] Answer: \answerNA{} %
    \item[] Justification: There is no research with human subjects in this work.

\item {\bf Declaration of LLM usage}
    \item[] Question: Does the paper describe the usage of LLMs if it is an important, original, or non-standard component of the core methods in this research? Note that if the LLM is used only for writing, editing, or formatting purposes and does \emph{not} impact the core methodology, scientific rigor, or originality of the research, declaration is not required.
    \item[] Answer: \answerYes{} %
    \item[] Justification: We describe the usage of LLMs if it is an important, original, or non-standard component of the core methods in this research.

\end{enumerate}

\end{document}